\documentclass[11pt]{article}

\usepackage[preprint]{acl}
\usepackage{amsfonts}
\usepackage{amsmath}
\usepackage{booktabs}
\usepackage{multirow}
\usepackage{pifont}

\usepackage{times}
\usepackage{latexsym}

\usepackage[T1]{fontenc}

\usepackage[utf8]{inputenc}

\usepackage{microtype}

\usepackage{inconsolata}

\usepackage{graphicx}
\graphicspath{{figures/}}
\usepackage{placeins}
\usepackage{tcolorbox}
\tcbuselibrary{breakable}


\title{\texttt{RouterKGQA}: Specialized--General Model Routing for Constraint-Aware Knowledge Graph Question Answering}

\author{
  Bo Yuan\textsuperscript{\rm 1}\thanks{Equal contribution.}~~~
  Hexuan Deng\textsuperscript{\rm 1,2}\footnotemark[1]~~~
  Xuebo Liu\textsuperscript{\rm 1}\thanks{Corresponding author.}~~~
  Min Zhang\textsuperscript{\rm 1}
  \\
  \textsuperscript{\rm 1}Institute of Computing and Intelligence, Harbin Institute of Technology, Shenzhen, China
  \\
  \textsuperscript{\rm 2}Zhongguancun Academy, Beijing, China
  \\
  \texttt{23s051006@stu.hit.edu.cn, hxuandeng@gmail.com}
  \\
  \texttt{\{liuxuebo,zhangmin2021\}@hit.edu.cn}
}

\begin{document}
\maketitle
\begin{abstract}
Knowledge graph question answering (KGQA) is a promising approach for mitigating LLM hallucination by grounding reasoning in structured and verifiable knowledge graphs. Existing approaches fall into two paradigms: \emph{retrieval-based} methods utilize small specialized models, which are efficient but often produce unreachable paths and miss implicit constraints, while \emph{agent-based} methods utilize large general models, which achieve stronger structural grounding at substantially higher cost. We propose \texttt{RouterKGQA}, a framework for specialized--general model collaboration, in which a specialized model generates reasoning paths and a general model performs KG-guided repair only when needed, improving performance at minimal cost. We further equip the specialized with constraint-aware answer filtering, which reduces redundant answers. In addition, we design a more efficient general agent workflow, further lowering inference cost. Experimental results show that RouterKGQA outperforms the previous best by 3.57 points in F1 and 0.49 points in Hits@1 on average across benchmarks, while requiring only 1.15 average LLM calls per question. Codes and models are available at~\url{https://github.com/Oldcircle/RouterKGQA}.
\end{abstract}

\section{Introduction}
\label{sec:introduction}

Recent advances in large language models (LLMs) have enabled remarkable progress across natural language processing tasks~\citep{achiam2023gpt4,grattafiori2024llama3}. However, LLMs remain prone to \textbf{factual hallucination}---generating plausible but factually incorrect content~\citep{wagner-etal-2025-mitigating}. A promising direction for mitigating hallucination is to ground LLM reasoning in knowledge graphs (KGs), which store large-scale factual knowledge in a structured, interpretable form~\citep{bollacker2008freebase,ji-etal-2024-retrieval,sui-etal-2025-fidelis}.

Current approaches to integrating LLMs with KGs fall into two paradigms. \textbf{Agent-based methods} utilize general LLMs to form an agent workflow that explores the KG through multi-round interactions to accumulate evidence~\citep{sun2024think-on-graph,chen2024plan-on-graph,dong-etal-2025-effiqa}. While effective, this incurs \emph{high computational overhead}, leading to significant inference latency and cost.

\textbf{Generation-based methods} train specialized LLMs to generate structured retrieval plans for KG execution~\citep{luo-etal-2024-chatkbqa,luo2024rog,pmlr-v267-luo25t}. While more efficient, they may generate \emph{hallucinated relations} absent from the target KG~\citep{tian-etal-2025-compkbqa}, especially under domain shift. Existing specialized LLMs also often overlook \emph{implicit constraints} in the question. For example, ``Who is the U.S. president?'' may correspond to multiple tail entities unless a temporal constraint is specified.

To address these issues, we propose \texttt{RouterKGQA}, a framework of specialized--general model collaboration, which achieves better performance with lower cost. Specifically, RouterKGQA dynamically routes queries between a specialized model, which is efficient and performs well in-domain, and a general model, which offers stronger generalization. We further improve both components. For the specialized model, we train it to capture implicit constraint semantics in questions, enabling more precise answer filtering. For the general agent workflow, we replace costly multi-round exploration with a cheaper exploration only on relations, which further reduces cost. As a result, \texttt{RouterKGQA} achieves strong accuracy while substantially lowering cost compared with conventional agent workflows. Our contributions are:

\begin{itemize}
    \item We propose RouterKGQA, a specialized--general routing framework that combines efficient specialized generation with general repair for KGQA.
    \item We improve both systems by introducing Constraint-aware Reasoning Paths (CRPs) for specialized models, and by designing a more efficient agent workflow for general models.
    \item Experiments on WebQSP and CWQ show that RouterKGQA achieves the highest F1 on both benchmarks with only 1.15 average LLM calls per question.
\end{itemize}

\section{Related Work}
\label{sec:related_work}

\paragraph{Factual Hallucination of LLMs.}
Despite strong language capabilities, LLMs remain prone to factual hallucination, producing fluent yet incorrect text~\citep{ji-etal-2023-hallucination-survey,huang-etal-2025-hallucination-survey}. Prompting-based methods such as chain-of-thought~\citep{wei-etal-2022-cot} and self-consistency~\citep{wang-etal-2023-selfconsistency} improve coherence but cannot guarantee factual correctness~\citep{ye-durrett-2022-unreliability}. Retrieval-augmented generation~\citep{lewis-etal-2020-rag,gao-etal-2024-rag-survey} grounds outputs in external evidence, but standard RAG retrieves unstructured text and lacks the relational structure needed for complex reasoning. KG-augmented approaches offer a more structured alternative by constraining reasoning to verifiable triples~\citep{pan-etal-2024-kgllm-survey,agrawal-etal-2024-kg-hallucination}.

\paragraph{Agent-based KGQA.}
Agent-based methods typically utilize general LLMs, treat the LLM as an interactive agent that explores the KG through multi-round tool calls. ToG~\citep{sun2024think-on-graph} performs iterative beam-search expansion over entities and relations. StructGPT~\citep{jiang-etal-2023-structgpt} provides specialized interfaces for structured data interaction. DoG~\citep{ma2025debate} introduces multi-role debate for question decomposition. PoG~\citep{chen2024plan-on-graph} decomposes questions into sub-objectives with backtracking and self-correction. While offering strong structural grounding, these methods require complex search over entities and relations and are often time-consuming. In contrast, we use a more efficient search scheme that operates only over relations, significantly reducing cost. We further route simple questions to a small specialized model, reducing cost even further.

\paragraph{Retrieval-based KGQA.}
These methods typically train specialized models to generate retrieval plans in a single pass, substantially reducing cost. \emph{Reasoning-path methods} generate relation paths for KG execution. RoG~\citep{luo2024rog} predicts relation sequences with a fine-tuned LLM and executes them via breadth-first search. GCR~\citep{pmlr-v267-luo25t} further constrains decoding with a KG-Trie to ensure path validity. GNN-RAG~\citep{mavromatis-karypis-2025-gnn} scores candidate entities with GNNs and extracts shortest paths as reasoning evidence. \emph{Logical-form methods} instead generate executable queries, such as S-expressions or SPARQL. ChatKBQA~\citep{luo-etal-2024-chatkbqa} generates S-expressions that are converted to SPARQL, while MemQ~\citep{xu-etal-2025-memory} reconstructs reasoning steps into query statements through tool calling. However, both paradigms struggle with implicit constraints within the question: path-based methods cannot explicitly encode them in specialized models, while logical-form methods handle them in nested formal languages that are harder to learn~\citep{tian-etal-2025-compkbqa}.

A central challenge of KGQA is \emph{relation hallucination}, i.e., generating relations absent from the target KG~\citep{tian-etal-2025-compkbqa}. Existing methods mitigate this mainly by restricting the search space, such as KG-Trie-constrained decoding~\citep{pmlr-v267-luo25t}, subgraph-based path extraction~\citep{mavromatis-karypis-2025-gnn}, and training-set relation libraries~\citep{xu-etal-2025-memory}. However, these strategies do not address the weak out-of-domain generalization of specialized models, which limits their performance upper bound. ChatKBQA~\citep{luo-etal-2024-chatkbqa} instead repairs erroneous relations via exhaustive post-hoc search, which is prohibitively slow for complex multi-hop queries.

Our method builds on reasoning-path methods and equips them with explicit constraint generation, improving accuracy. To further address OOD-induced hallucination, we route to a general model when specialized models fail to handle the question.

\section{Method}
\label{sec:method}

\subsection{Problem Formulation}
\label{sec:irc_overview}

A knowledge graph is defined as
\begin{equation}
\mathbb{G}=\{(s,r,o)\mid s\in\mathbb{E},\ r\in\mathbb{R},\ o\in\mathbb{E}\cup\mathbb{L}\},
\label{eq:kg_def}
\end{equation}
where $\mathbb{E}$, $\mathbb{R}$, and $\mathbb{L}$ denote the sets of entities, relations, and literal values, respectively. Given an $N$-hop question $q$ over $\mathbb{G}$, the goal is to recover the reasoning path
\begin{equation}
\mathbf{P}=\langle (s_1,r_1,o_1),\ldots,(s_N,r_N,o_N)\rangle,
\label{eq:path_def}
\end{equation}
where $o_N$ is the final answer to $q$, and adjacent triples satisfy $s_{i+1}=o_i$ for $i=1,\ldots,N-1$.

\begin{figure*}[t]
\centering
\includegraphics[width=0.9\textwidth]{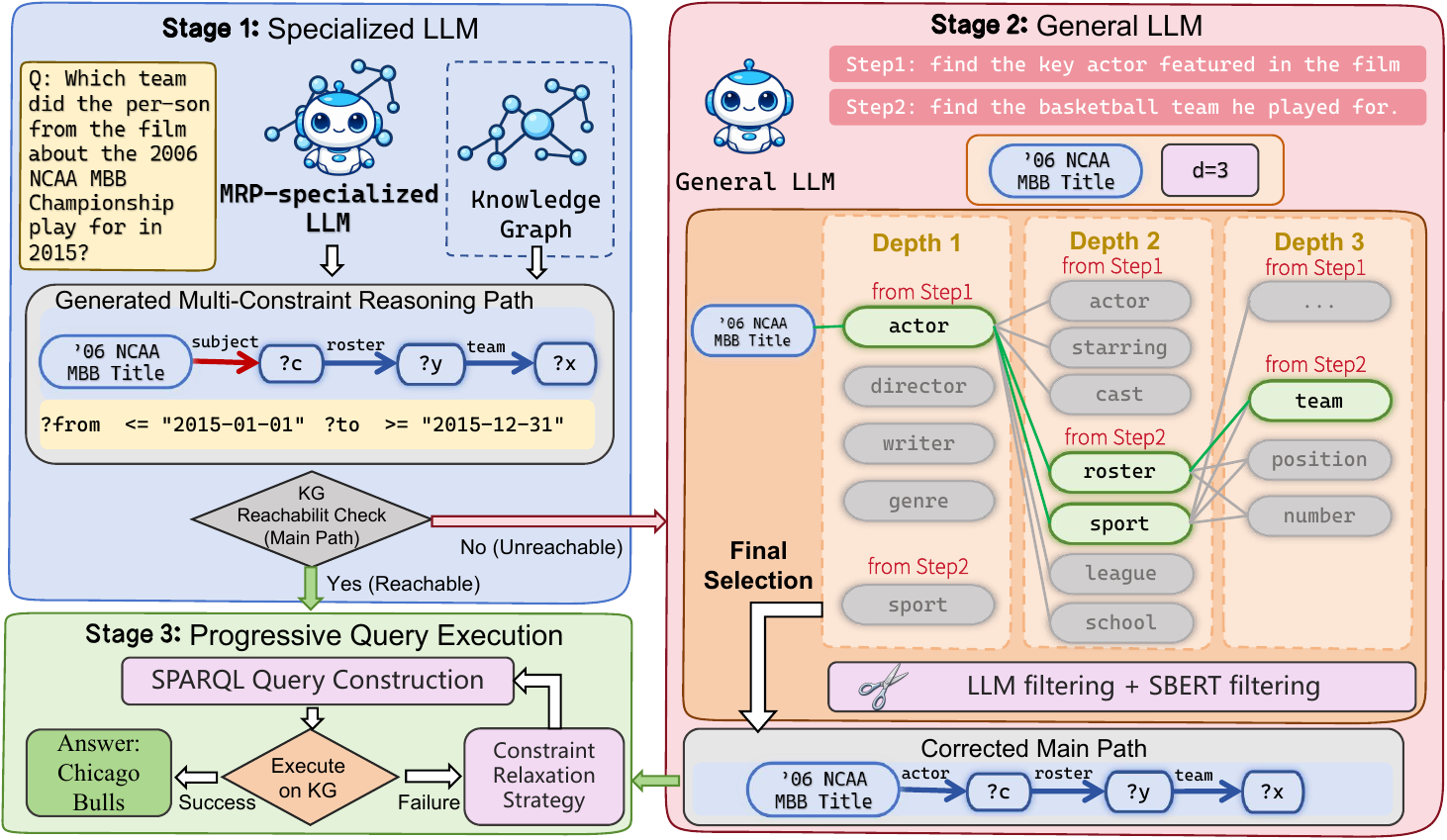}
\caption{Overview of the RouterKGQA framework.}
\label{fig:method}
\end{figure*}

To solve this problem, we propose \texttt{RouterKGQA}, as illustrated in Figure~\ref{fig:method}. The framework consists of three stages. First, we train a specialized model, which maps a natural-language question to a \emph{Constraint-aware Reasoning Path} (CRP), which contains a main path and its associated constraints (\S\ref{sec:mrp_generation}). Second, if the main path is unreachable, an agent-based workflow repairs it via beam search, which requires fewer LLM calls than other methods (\S\ref{sec:kg_correction}). Finally, the corrected CRP is converted into a SPARQL query to retrieve the final answer (\S\ref{sec:progressive_exec}). The prompts we used are in Appendix~\ref{app:prompts}.

\subsection{Constraint-Aware Path Generation with Specialized LLMs}
\label{sec:mrp_generation}

We describe how the specialized model generates a CRP from a natural-language question, and reachability verification that determines whether repair is needed (Stage 1).

\paragraph{Initial entity.}
A CRP starts from a topic entity name $s_1$. We let the specialized model first generate the initial entity, then ground this predicted name to its corresponding entity using ELQ~\citep{li2020efficient}, yielding the grounded topic entity $s_1 \in \mathbb{E}$.

\paragraph{Main path.}
Following \citet{luo2024rog}, we represent the reasoning path as a flat sequence of relations rather than as a nested logical form. This format is easier for small LLMs to learn because it avoids recursive syntax and function tokens, and it keeps the main path separate from constraints, enabling independent reachability verification and progressive constraint relaxation. Specifically, from $\mathbf{P}$ we extract the relation sequence $\mathbf{R}=\langle r_1,r_2,\ldots,r_N\rangle$ as the main path. Starting from $s_1$, we follow the relations in $\mathbf{R}$ hop by hop to reach the answer $o_N$.

\paragraph{Constraint set.}
However, a relation path alone is often insufficient because, in a KG, a pair $(s,r)$ may correspond to multiple objects, yielding multiple candidate answers. To resolve this ambiguity, building on prior work, we equip the specialized model with the ability to generate constraints. We generate constraints over intermediate entities, which is defined as $C=(i,r_i^c,o_i^c)$, requiring $o_i$ to satisfy relation $r_i^c$ with value or condition $o_i^c$. Formally, this restricts $o_i$ to
\begin{equation}
\mathbb{E}_i^c=\{e\in\mathbb{E}\mid (e,r_i^c,o_i^c)\in\mathbb{G}\}.
\end{equation}
For example, for the question ``Which U.S.\ president graduated from Harvard and assumed office after 2000?'', the main path $\mathbf{R}=\langle\text{country.presidents},\ \text{president.office\_holder}\rangle$ retrieves candidate presidents. We then generate $C=(2,\ \text{education.institution},\ \text{Harvard})$ to require the candidate to be linked to \text{Harvard}, and $C=(2,\ \text{position.from},\ {\geq}\text{``2000''})$ to retain only those whose term started in or after 2000.

\paragraph{Specialized Model Training.}
To construct instruction fine-tuning data, we convert the SPARQL annotations in the training set into gold CRPs using the deterministic procedure described in Appendix~\ref{app:mrp_conversion}. Each training instance pairs a natural-language question as input with the processed gold CRP as output, following the instruction template in Appendix~\ref{app:prompt_mrp}. We fine-tune the specialized model with LoRA~\citep{hu2022lora}; implementation details are provided in Section~\ref{sec:exp_setup}.

\paragraph{Reachability verification.}
The generated main path may still be inconsistent with the KG because of relation errors, direction mismatches, or structural mistakes. We therefore execute the skeleton query induced by $\mathbf{R}$ on $\mathbb{G}$. If it returns a non-empty result set, the main path is reachable. Otherwise, we assume that the specialized model has failed on this case and invoke the general model.

\subsection{Efficient Path Repair with General LLMs}
\label{sec:kg_correction}

The specialized model is accurate and cost-effective on in-domain questions, but often fails under domain shift. When reachability verification fails, we treat the case as beyond the capability of the specialized model and route the question to a stronger agent workflow (Stage 2). To keep this stage efficient, we avoid iterative search over entities and relations. Instead, we first ask the general model to generate a sequence of high-level subtasks in one shot, and then recover an executable relation path via KG-guided beam search.

\paragraph{Reasoning blueprint generation.}
Given a question $q$, the general model decomposes it into a sequence of textual subtasks
\begin{equation}
\mathbf{T}=\langle t_1,t_2,\ldots,t_M\rangle,
\end{equation}
where each $t_k$ describes one semantic step in the reasoning process. These blueprints provide semantic guidance for search, but are not assumed to align one-to-one with KG relations.

\paragraph{Beam search over relation paths.}
To map the above subtasks to a relation path, we search over relation sequences rather than joint entity--relation states. Let
\begin{equation}
\mathcal{B}_{i-1}=\{\mathbf{R}_{i-1,1},\ldots,\mathbf{R}_{i-1,w}\}
\end{equation}
denote the beam at depth $i-1$, where $w$ is the beam width and
\begin{equation}
\mathbf{R}_{i-1,j}=\langle r_{1,j},r_{2,j},\ldots,r_{i-1,j}\rangle
\end{equation}
is the $j$-th partial relation path.

For each partial path $\mathbf{R}_{i-1,j}$, we first compute the set of entities reachable from the topic entity $s_1$:
\begin{equation}
\mathcal{E}_{i-1,j}=\operatorname{Reach}(s_1,\mathbf{R}_{i-1,j}),
\label{eq:reachable_entities}
\end{equation}
where $\operatorname{Reach}(s_1,\mathbf{R})$ denotes the set of entities reached by following the relation sequence $\mathbf{R}$ from $s_1$ on $\mathbb{G}$. For initialization, $\mathcal{E}_{0,1}=\{s_1\}$. We then enumerate all outgoing relations from these reachable entities:
\begin{equation}
\mathcal{R}^{\mathrm{next}}_{i,j}
=
\{\,r \mid \exists e\in \mathcal{E}_{i-1,j},\ \exists o,\ (e,r,o)\in\mathbb{G}\,\}.
\label{eq:next_relations}
\end{equation}

\paragraph{Relation selection.}
We first perform coarse filtering to retain relations that are most relevant to the blueprint. Because blueprint steps and KG relations are not always perfectly aligned, we score each candidate relation against every blueprint $t_k$ using sentence-BERT (SBERT; \citealp{reimers-gurevych-2019-sentence}) and keep the top-$x$ relations:
\begin{equation}
\widetilde{\mathcal{R}}_{i,j,k}
=
\operatorname{Top}_x
\bigl(
\mathcal{R}^{\mathrm{next}}_{i,j};
\operatorname{sim}(t_k,r)
\bigr),
\label{eq:relation_filter}
\end{equation}
where $\operatorname{Top}_x(\mathcal{S}; f(r))$ denotes the subset of $\mathcal{S}$ containing the $x$ elements with the largest values of $f(r)$, where $r \in \mathcal{S}$. $\operatorname{sim}(t_k,r)$ is the cosine similarity between the SBERT embeddings of $t_k$ and $r$.

We then expand each partial path with the filtered relations and obtain the candidate set
\begin{equation}
\widetilde{\mathcal{P}}_i
=
\bigcup_{j=1}^{w}
\{\,\mathbf{R}_{i-1,j}\oplus r \mid \exists k,\ r\in\widetilde{\mathcal{R}}_{i,j,k}\,\},
\label{eq:expanded_paths}
\end{equation}
where $\oplus$ appends a relation to a path.

\paragraph{Path selection.}
To further reduce the search space, we score each expanded path against the full question $q$ and retain the top-$y$ candidates:
\begin{equation}
\widehat{\mathcal{P}}_i
=
\operatorname{Top}_y(\widetilde{\mathcal{P}}_i; \operatorname{sim}(q,\mathbf{R})),
\label{eq:path_filter}
\end{equation}
where each path $\mathbf{R}$ is linearized by concatenating its relation names.

Finally, the general model further reduces the candidate set by selecting the best $w$ paths from $\widehat{\mathcal{P}}_i$, forming the next beam $\mathcal{B}_i$. Each search step therefore requires only one LLM call, while the remaining filtering is handled by small SBERT at negligible cost. Including the initial blueprint generation, the total number of LLM calls is at most $d+1$, where $d$ is the maximum search depth. Moreover, because every expansion is derived from KG-reachable entities, all candidate paths in the beam remain executable.

\paragraph{Stopping criterion.}
To determine when to stop the search, we use the hop count predicted by the specialized model as the maximum search depth $d$. Even when the specialized model predicts an incorrect relation path, its depth prediction remains highly reliable: with Llama-2-7B, the predicted depth matches the ground-truth depth in 97.20\% of cases on CWQ and 98.70\% on WebQSP. After depth $d$, we reduce the beam size to 1 and return the remaining path as the corrected main path $\mathbf{R}$.

\subsection{Progressive Query Execution}
\label{sec:progressive_exec}

In this section, we explain how the corrected CRP is converted into an executable SPARQL query and retrieve answers from the KG. We also describe the progressive constraint relaxation strategy for improving robustness when full execution fails (Stage 3).

\paragraph{Execution with full constraints.}
After correction, we deterministically convert the CRP into an executable SPARQL query. The main path $\mathbf{R}=\langle r_1,\ldots,r_N\rangle$ is mapped to a chain of triple patterns rooted at the topic entity $s_1$, with each $r_i$ corresponding to one hop. Each constraint $C=(i,r_i^c,o_i^c)$ is then translated into an additional condition: entity constraints are converted into triple patterns requiring the intermediate entity $o_i$ to connect to $o_i^c$ via $r_i^c$, numeric constraints are converted into \texttt{FILTER} clauses imposing inequality conditions on numeric or temporal attributes, and string constraints, when present, are converted into literal-matching \texttt{FILTER} clauses. Executing the resulting query yields the answer set
\begin{equation*}
\mathcal{A} \leftarrow \operatorname{Execute}(\operatorname{ConvertCRP}(\operatorname{CRP}), \mathbb{G}).
\end{equation*}
If execution succeeds, we return $\mathcal{A}$ directly. Details on constructing gold CRPs from SPARQL annotations for training are provided in Appendix~\ref{app:mrp_conversion}.

\paragraph{Constraint relaxation.}
If execution with all constraints fails, e.g., due to noisy constraint predictions or KG incompleteness, we progressively relax lower-priority constraints while keeping the corrected main path fixed, and re-execute the simplified query. This strategy improves robustness without discarding the core reasoning chain.

\begin{table*}[t]
\centering
\small
\scalebox{0.95}{
\begin{tabular}{llcccc}
\toprule
\textbf{Category} & \textbf{Method}
& \multicolumn{2}{c}{\textbf{WebQSP}}
& \multicolumn{2}{c}{\textbf{CWQ}} \\
\cmidrule(lr){3-4} \cmidrule(lr){5-6}
 & & Hits@1 & F1 & Hits@1 & F1 \\
\midrule[1.5pt]
LLM-only & Llama-2-7B \citep{touvron2023llama2} & 53.87 & 35.67 & 22.74 & 18.11 \\
 & Llama-3.1-8B \citep{grattafiori2024llama3} & 55.96 & 31.61 & 30.81 & 21.83 \\
 & GPT-4o-mini & 64.86 & 48.59 & 40.89 & 40.89 \\
\midrule
IR-based & KV-Mem \citep{miller-etal-2016-kvmem} & 46.70 & 34.50 & 21.10 & 15.70 \\
 & PullNet \citep{sun2019pullnet} & 68.10 & -- & 47.20 & -- \\
 & EmbedKGQA \citep{sun2020embedkgqa} & 66.60 & -- & 44.70 & -- \\
 & NSM+h \citep{he-etal-2021-nsm} & 74.30 & 67.40 & 48.80 & 44.00 \\
 & TransferNet \citep{shi-etal-2021-transfernet} & 71.40 & -- & 48.60 & -- \\
 & Subgraph Retrieval \citep{sun-etal-2020-subgraph-retrieval} & 69.50 & 64.10 & 50.20 & 47.10 \\
\midrule
LLM-based
 & ToG (GPT-4o-mini) \citep{sun2024think-on-graph} & 77.20 & 50.04 & 60.67 & 41.90 \\
 & PoG (GPT-4o-mini) \citep{chen2024plan-on-graph} & 83.13 & 61.15 & 64.67 & 55.56 \\
 & InteractiveKBQA (GPT-4) \citep{xiong-etal-2024-interactive-kbqa} & 72.47 & -- & 59.17 & -- \\
 & EffiQA \citep{dong-etal-2025-effiqa} & 82.90 & -- & 69.50 & -- \\
 & RoG (Llama-2-7B) \citep{luo2024rog} & 86.18 & 70.24 & 61.71 & 54.98 \\
 & ChatKBQA (Llama-2-7B) \citep{luo-etal-2024-chatkbqa} & 86.33 & 82.13 & 85.16 & 80.94 \\
 & MemQ \citep{xu-etal-2025-memory} & 88.74 & 77.90 & 88.16 & 75.56 \\
 & GNN-RAG \citep{mavromatis-karypis-2025-gnn} & 90.73 & 73.49 & 68.65 & 60.45 \\
 & GCR (Llama-3.1-8B + GPT-4o-mini) \citep{pmlr-v267-luo25t} & 91.68 & 75.27 & 74.64 & 62.94 \\
\midrule
\textbf{Ours} & RouterKGQA (Llama-2-7B + GPT-4o-mini) & 90.73 & \textbf{86.43} & 88.96 & 83.72 \\
 & RouterKGQA (Llama-3.1-8B + GPT-4o-mini) & 91.15 & 86.38 & \textbf{89.66} & \textbf{83.78} \\
\bottomrule
\end{tabular}}
\caption{Performance comparison on WebQSP and CWQ (\%). Results for methods without public code are taken from the original papers; all other results are reproduced in our environment using the same Freebase snapshot. See Appendix~\ref{app:metric_details} for unified metric definitions and Appendix~\ref{app:implementation} for reproduction settings and implementation details.}
\label{tab:kbqa_performance_simple}
\end{table*}

\begin{table*}[t]
\centering
\small
\scalebox{0.95}{
\begin{tabular}{llccccc}
\toprule
\textbf{Category} & \textbf{Method} &
\textbf{Hits@1} & \textbf{F1} &
\begin{tabular}[c]{@{}c@{}}\textbf{Avg. \#}\\\textbf{LLM Calls}\end{tabular} &
\begin{tabular}[c]{@{}c@{}}\textbf{Avg. \#}\\\textbf{Tokens}\end{tabular} &
\begin{tabular}[c]{@{}c@{}}\textbf{Avg. Cost}\\\textbf{(USD) / 10K Qs}\end{tabular} \\
\midrule
\multirow{5}{*}{\textbf{Retrieval-based}}
& GNN-RAG  & 90.73 & 73.49 & 2.00 & 778.17 & 0.56 \\
& RoG      & 86.18 & 70.24 & 2.00 & 592.55 & 0.45 \\
& ChatKBQA & 86.33 & 82.13 & 1.00 & 208.08 & 0.46 \\
& GCR      & 91.68 & 75.27 & 2.00 & 725.57 & 0.89 \\
& MemQ     & 88.74 & 77.90 & 1.00 & 175.62 & 0.21 \\
\midrule
\multirow{2}{*}{\textbf{Agent-based}}
& ToG  & 77.20 & 50.04 & 10.91 & 8{,}426.59 & 19.35 \\
& PoG  & 83.13 & 61.15 & 7.64 & 4{,}704.47 & 8.07 \\
\midrule
\multirow{2}{*}{\textbf{Ours}}
& RouterKGQA (Stage~2+3) & 86.07 & 70.68 & 3.31 & 1{,}353.78 & 2.57 \\
& RouterKGQA (Full) & 91.15 & \textbf{86.38} & 1.15 & 612.68 & 0.57 \\
\bottomrule
\end{tabular}
}
\caption{Efficiency and performance comparison on WebQSP (\%). RouterKGQA (Full) achieves the best F1 at a cost comparable to retrieval-based methods, while RouterKGQA (Stage~2+3), i.e., the agent-based-only configuration, remains substantially cheaper than representative agent-based baselines. Costs are estimated using Llama-2-7B pricing (\$0.05/1M input, \$0.25/1M output) and Llama-3.1-8B pricing (\$0.10/1M input, \$0.10/1M output) from Artificial Analysis, and official GPT-4o-mini pricing (\$0.15/1M input, \$0.60/1M output) from OpenAI, all accessed 2026-02-14.}
\label{tab:efficiency_webqsp}
\end{table*}

\section{Experiments}
\label{sec:experiments}

\subsection{Experimental Setup}
\label{sec:exp_setup}

\paragraph{Datasets.}
We evaluate on two widely used KBQA benchmarks: \textbf{WebQuestionsSP (WebQSP)}~\citep{yih-etal-2016-value} and \textbf{Complex WebQuestions (CWQ)}~\citep{talmor-berant-2018-web}, using Freebase~\citep{bollacker2008freebase} as the underlying KG. Compared with WebQSP, CWQ generally contains questions with greater hop depth and more complex constraints, and is therefore substantially more challenging. More detailed dataset statistics and descriptions are provided in Appendix~\ref{app:dataset_details}.

\paragraph{Metrics and baselines.}
Following prior work, we report Hits@1 and F1. Hits@1 measures the proportion of questions for which the predicted answer set overlaps with the gold answer set, while F1 is the mean F1 score between the predicted and gold answer sets. We compare against three categories: (1) LLM-only methods; (2) IR-based methods using subgraph retrieval; and (3) LLM-based methods, including agent-based and retrieval-based approaches. Detailed metric definitions are provided in Appendix~\ref{app:metric_details}.

\paragraph{Configuration of RouterKGQA.}
RouterKGQA uses a LoRA-tuned Llama-2-7B~\citep{touvron2023llama2} or Llama-3.1-8B~\citep{grattafiori2024llama3} specialized model for CRP generation, and GPT-4o-mini ($T{=}0$) for path repair. The specialized model is trained on (question, gold CRP) pairs derived from SPARQL annotations. Unless otherwise noted, we use the same optimization settings for the specialized model, including a learning rate of $5\times10^{-5}$, cosine scheduler, batch size 4, gradient accumulation 4, and bf16. For Stage~2, we set the relation filter size to $x{=}4$, the path filter size to $y{=}10$, and the beam width to $w{=}3$, and use SBERT (nomic-embed-text-v1; \citealp{za-nomic-2025}) for semantic matching. In Stage~3, constraints are relaxed in the order string $\rightarrow$ numeric $\rightarrow$ entity. All experiments use seed 42. Additional implementation and reproduction details are provided in Appendix~\ref{app:implementation}.


\subsection{Main Results}
\label{sec:main_results}

\paragraph{RouterKGQA improves F1 while preserving strong Hits@1.}
Table~\ref{tab:kbqa_performance_simple} shows that RouterKGQA consistently improves F1 without sacrificing Hits@1, indicating that constraint-aware execution effectively filters spurious candidates. Compared with the strongest prior F1 baseline, RouterKGQA improves F1 by 4.30 points on WebQSP and 2.84 points on CWQ. At the same time, its Hits@1 remains comparable to the strongest prior baseline on WebQSP and surpasses it on CWQ, showing that the F1 gains come from better candidate filtering while preserving the correct answer in most cases.

\paragraph{Agent-based methods are often more effective, but substantially more expensive.}
Table~\ref{tab:efficiency_webqsp} shows that representative agent-based methods, such as ToG and PoG, outperform simple LLM-only baselines on average, reflecting the benefit of explicit multi-step KG exploration. However, this gain comes at a substantial efficiency cost: compared with retrieval-based methods, agent-based methods require 5.80$\times$ more LLM calls and incur 26.67$\times$ higher cost on average.

\paragraph{Our agent-based path repair is much cheaper, and model routing reduces the cost further.}
RouterKGQA (Stage~2+3), i.e., the agent-based-only configuration, already outperforms representative agent-based baselines while reducing cost by 3.14$\times$ relative to PoG and 7.53$\times$ relative to ToG. Adding model routing further improves the trade-off: compared with the agent-based-only setting, RouterKGQA (Full) reduces cost by 4.51$\times$ and LLM calls by 2.88$\times$, while improving F1 by 15.70 points. This shows that routing between retrieval-based generation and agent-based repair is more effective than relying on either strategy alone.

\subsection{Ablation Study}
\label{sec:ablation}

We examine the effects of model routing, stopping criteria, constraint modeling, and representation design.

\begin{table}[t]
\centering
\small
\setlength{\tabcolsep}{3.5pt}
\scalebox{0.95}{
\begin{tabular}{lcccc}
\toprule
\textbf{Variants} & \multicolumn{2}{c}{\textbf{WebQSP}} & \multicolumn{2}{c}{\textbf{CWQ}} \\
\cmidrule(lr){2-3}\cmidrule(lr){4-5}
& \textbf{Hits@1} & \textbf{F1} & \textbf{Hits@1} & \textbf{F1} \\
\midrule
RouterKGQA & \textbf{91.15} & \textbf{86.38} & \textbf{89.66} & \textbf{83.78} \\
\midrule
\multicolumn{5}{l}{\emph{Stage-level ablations}} \\
w/o Stage~1 (Stage~2+3) & 86.07 & 70.68 & 70.67 & 59.30 \\
w/o Stage~2 (Stage~1+3) & 89.02 & 84.68 & 88.05 & 82.12 \\
\midrule
\multicolumn{5}{l}{\emph{Stopping criterion variants}} \\
w/ blueprint step count & 88.47 & 83.96 & 88.59 & 82.81 \\
w/ LLM self-judge & 90.76 & 86.11 & 89.44 & 83.40 \\
\midrule
\multicolumn{5}{l}{\emph{Constraint ablations (on Stage~1+3)}} \\
w/o Entity Constraint & 88.65 & 78.21 & 87.20 & 58.32 \\
w/o Numeric Constraint & 89.62 & 81.66 & 88.67 & 71.15 \\
w/o String Constraint & 89.02 & 84.58 & 87.96 & 80.72 \\
w/o All Constraints & 89.45 & 75.56 & 88.28 & 46.60 \\
\bottomrule
\end{tabular}
}
\caption{Ablation study of RouterKGQA (\%). Constraint ablations are conducted on Stage~1+3. Stage~1: retrieval-based CRP generation; Stage~2: agent-based KG-guided repair; Stage~3: progressive query execution.}
\label{tab:ablation}
\end{table}

\paragraph{Agent-based repair brings marginal gains.}
Table~\ref{tab:ablation} shows that Stage~1 is the primary source of performance, while Stage~2 provides reliable additional improvements. Removing Stage~2 leads to smaller drops than stage 1 on both benchmarks, but the drop is consistent. This shows that agent-based path repair delivers meaningful marginal gains by correcting unreachable relation paths that the specialized model fails to recover.

\paragraph{Depth prediction is robust.}
In the main experiments, we use the hop count predicted by the specialized model as the stopping criterion. We compare it with two alternative variants as stopping criterion. Using blueprint step count consistently underperforms, indicating that semantic subtasks do not align perfectly with KG hop structure. We further let the general LLM self-judge when to stop. This variant performs closer to the default setting, but is still slightly worse and requires an additional LLM call. These results support using the specialized model to control the depth of agent-based search, achieving the best performance with minimal cost.

\paragraph{Constraints mainly improve precision.}
Within Stage~1+3, removing all constraints leaves Hits@1 nearly unchanged but substantially reduces F1. This shows that constraints primarily improve precision by filtering spurious candidates rather than changing top-1 reachability. Among the constraint types, entity constraints contribute the most, followed by numeric constraints.

To further verify the effectiveness of our constraints, we apply them to the baseline GCR, with results shown in Table~\ref{tab:gcr_ablation}. The results show that removing GCR filtering yields relatively high Hits@1 but much lower F1, while adding our generated constraint filtering further strengthens this effect, especially on the harder CWQ benchmark. This suggests that its filtering mechanism remains insufficient. Besides, even the best GCR variant remains clearly below RouterKGQA, suggesting that our explicit CRP formulation is more effective than post-hoc filtering over monolithic outputs.

\begin{table}[t]
\centering
\small
\setlength{\tabcolsep}{2pt}
\scalebox{0.95}{
\begin{tabular}{lcccc}
\toprule
\textbf{Variants} & \multicolumn{2}{c}{\textbf{WebQSP}} & \multicolumn{2}{c}{\textbf{CWQ}} \\
\cmidrule(lr){2-3}\cmidrule(lr){4-5}
& \textbf{Hits@1} & \textbf{F1} & \textbf{Hits@1} & \textbf{F1} \\
\midrule
GCR (w/o LLM Filter) & 92.13 & 57.00 & 74.99 & 49.65 \\
GCR & 91.68 & 75.27 & 74.64 & 62.94 \\
GCR (w/ RouterKGQA Filter) & 87.13 & 83.95 & 80.46 & 68.05 \\
\midrule
RouterKGQA & \textbf{91.15} & \textbf{86.38} & \textbf{89.66} & \textbf{83.78} \\
\bottomrule
\end{tabular}
}
\caption{Reference comparison with GCR variants (\%). RouterKGQA is included for reference.}
\label{tab:gcr_ablation}
\end{table}

\begin{table}[t]
\centering
\small
\setlength{\tabcolsep}{3pt}
\scalebox{0.95}{
\begin{tabular}{llcc}
\toprule
\textbf{Dataset} & \textbf{Metric} & \textbf{ChatKBQA} & \textbf{RouterKGQA} \\
\midrule
\multirow{2}{*}{WebQSP} & Skeleton Accuracy & 82.43 & \textbf{85.78} \\
& Exact Match & 62.42 & \textbf{69.68} \\
\midrule
\multirow{2}{*}{CWQ} & Skeleton Accuracy & 74.00 & \textbf{75.62} \\
& Exact Match & 54.23 & \textbf{54.63} \\
\bottomrule
\end{tabular}
}
\caption{First-prediction generation accuracy (\%) of ChatKBQA (S-expression) and RouterKGQA (CRP), both using Llama-2-7B.}
\label{tab:mrp_vs_sexpr}
\end{table}

\paragraph{CRPs are easier to generate.}
We further compare whether CRP generation offers intrinsic advantages over the S-expression format, e.g., ChatKBQA~\citep{luo-etal-2024-chatkbqa}. Using the same backbone (Llama-2-7B), we measure: (1)~\textbf{skeleton accuracy}---whether the predicted structure matches the gold, ignoring entity/literal values; and (2)~\textbf{exact match}---whether the prediction is fully correct.

As shown in Table~\ref{tab:mrp_vs_sexpr}, CRP achieves higher skeleton accuracy on both benchmarks (+3.35\% on WebQSP, +1.62\% on CWQ), indicating that the flat triple-sequence format is structurally easier for a 7B model to learn than nested S-expressions with function keywords. The advantage is more pronounced on exact match (+7.26\% on WebQSP), confirming that CRP's explicit constraint separation also improves value-level prediction.

\subsection{Analysis by Query Complexity}
\label{sec:complexity_analysis}

\begin{table}[t]
\centering
\small
\setlength{\tabcolsep}{2.3pt}
\scalebox{0.95}{
\begin{tabular}{llccccc}
\toprule
\textbf{Dimension} & \textbf{Category} & \textbf{N}
& \textbf{Full} & \textbf{S1+S3} & \textbf{$\Delta$F1}
& \textbf{GCR} \\
\midrule
\multirow[c]{4}{*}{\shortstack[l]{Composition\\Type}}
 & Conjunction  & 1575 & \textbf{85.95} & 83.18 & +2.77 & 67.52 \\
 & Composition  & 1546 & \textbf{82.34} & 81.91 & +0.43 & 62.39 \\
 & Comparative  & 213  & \textbf{82.02} & 80.88 & +1.14 & 49.62 \\
 & Superlative  & 197  & \textbf{79.70} & 76.53 & +3.17 & 45.05 \\
\midrule
\multirow[c]{4}{*}{\shortstack[l]{Constraint\\Complexity}}
 & 0  & 1176 & \textbf{85.26} & 84.70 & +0.56 & 65.33 \\
 & 1  & 1511 & \textbf{84.90} & 82.94 & +1.96 & 61.03 \\
 & 2  & 633  & \textbf{78.72} & 75.76 & +2.96 & 63.73 \\
 & 3+ & 211  & \textbf{82.76} & 80.88 & +1.88 & 61.04 \\
\bottomrule
\end{tabular}
}
\caption{CWQ performance by query complexity (F1, \%). $\Delta$F1 = Full $-$ Stage~1+3 (i.e., the contribution of Stage~2 correction). GCR results are included for cross-method comparison.}
\label{tab:complexity_analysis}
\end{table}

Table~\ref{tab:complexity_analysis} jointly analyzes RouterKGQA's performance across two complexity dimensions on CWQ.

\paragraph{Advantage over GCR across composition types.}
RouterKGQA achieves consistent and large improvements over GCR across all four CWQ composition types. The advantage is most pronounced on \textbf{comparative} (+32.40 F1) and \textbf{superlative} (+34.65 F1) queries---precisely the types that demand explicit numeric and ordering constraints, where GCR's constraint-free path generation is most inadequate. Even on simpler conjunction and composition types, RouterKGQA maintains substantial margins (+18.43 and +19.95 F1), demonstrating broad improvements across all query structures.

\paragraph{Stage~2 contribution amplified by constraints.}
Stage~2's correction contributes most on \textbf{superlative} queries ($\Delta$F1 = +3.17), where precise paths are essential for executing ordering constraints, and on \textbf{conjunction} queries ($\Delta$F1 = +2.77), which involve multiple joined conditions. The breakdown by constraint complexity reveals a clear trend: the gain is largest on questions of moderate difficulty, i.e., those with 2 constraints. For overly simple questions, the specialized model already handles them well, while for overly difficult questions, the repair success rate also decreases. This analysis further provides insight into when Stage~2 repair is most effective.

\subsection{Specialized--General Model Analysis}
\label{sec:model_type}

\begin{table}[t]
\centering
\small
\setlength{\tabcolsep}{3pt}
\scalebox{0.95}{%
\begin{tabular}{llcccc}
\toprule
\multirow{2}{*}{\textbf{Role}} & \multirow{2}{*}{\textbf{Model}}
& \multicolumn{2}{c}{\textbf{WebQSP}} & \multicolumn{2}{c}{\textbf{CWQ}} \\
\cmidrule(lr){3-4}\cmidrule(lr){5-6}
& & \textbf{Hits@1} & \textbf{F1} & \textbf{Hits@1} & \textbf{F1} \\
\midrule
\multirow[c]{4}{*}{\shortstack[l]{Specialized\\(Stage 1)}}
& Qwen2.5-7B       & 88.04 & 84.67 & 88.19 & 81.32 \\
& Qwen3-8B       & 88.04 & 84.80 & 89.32 & 83.32 \\
& Llama-2-7B     & 90.73 & \textbf{86.43} & 88.96 & 83.72 \\
& Llama-3.1-8B   & \textbf{91.15} & 86.38 & \textbf{89.66} & \textbf{83.78} \\
\midrule
\multirow[c]{5}{*}{\shortstack[l]{General\\(Stage 2)}}
& Qwen-2.5-7B      & 90.79 & 86.25 & 89.49 & 83.63 \\
& Llama-2-7B     & 89.63 & 85.30  & 88.25 & 82.67 \\
& Llama-3.1-8B   & 89.57 & \textbf{86.90} & 89.47 & 83.35 \\
& ChatGPT        & 89.38 & 85.20 & 89.21 & 83.49 \\
& GPT-4o-mini   & \textbf{91.15} & 86.38 & \textbf{89.66} & \textbf{83.78} \\
\bottomrule
\end{tabular}%
}
\caption{Impact of model choice in RouterKGQA's specialized and general roles.}
\label{tab:irc_llm_webqsp_cwq}
\end{table}

Table~\ref{tab:irc_llm_webqsp_cwq} examines the impact of model choice in each role. When varying the specialized model (fixing GPT-4o-mini as the general model), stronger models produce substantially better results: Llama-3.1-8B outperforms Qwen2.5-7B by +3.11 Hits@1 on WebQSP and +2.46 F1 on CWQ. This confirms that CRP generation quality---which benefits from fine-tuning on KG-structured data---is the primary driver of end-to-end performance.

When varying the general model (fixing Llama-3.1-8B as specialized), the gap between models is notably smaller. Even Llama-2-7B as the general achieves 89.63\% Hits@1 on WebQSP, a gap of only 1.52 points with GPT-4o-mini. Among stronger generals, performance differences are within 1.8 points. This is because RouterKGQA's correction task is relatively simple---selecting the best path from a small candidate set---rather than generating answers from scratch, making RouterKGQA deployable with fully open-source models at minimal performance loss. Appendix~\ref{app:subq_consistency} further shows that RouterKGQA yields reasoning paths that are more semantically consistent with the original question.

\section{Conclusion}
\label{sec:conclusion}

We proposed RouterKGQA, a framework for knowledge graph question answering based on specialized--general model routing, which combines efficient specialized generation with general model repair. We further improve both components: for the specialized model, we introduce Constraint-aware Reasoning Paths for deterministic answer filtering. For the general workflow, we design an efficient KG-guided repair procedure that reduces the cost of multi-round exploration. Experiments on WebQSP and CWQ show that RouterKGQA achieves strong performance gains, while requiring only 1.15 average LLM calls per question. Further analysis shows that the repair stage provides consistent gains at low cost, and that the overall framework remains effective across different query types and model choices.

\section*{Limitations}

RouterKGQA has several limitations.
First, the current CRP representation supports only flat entity, numeric,
and string constraints. This keeps the representation simple and easier
for specialized models to learn, but does not explicitly capture more
nested or compositional constraint structures.
Second, the current routing decision uses main-path reachability as a
simple and efficient signal. However, a reachable path may still be
semantically incorrect. Incorporating finer-grained signals, such as
generation confidence or semantic consistency, could further improve
robustness.
Third, like other KGQA pipelines, our framework benefits from accurate
entity linking and a sufficiently complete underlying KG; errors in
topic-entity disambiguation or missing relations or literals may still
affect both correction and final answer filtering.
Fourth, our experiments are limited to English benchmarks on Freebase
(WebQSP and CWQ). Although the framework is general, its transferability
to other knowledge graphs, domains, or multilingual settings remains to
be validated.
Finally, although progressive constraint relaxation improves robustness
when full query execution fails, it may occasionally return broader
answer sets than desired when important constraints are removed.

\section*{Ethical considerations}
Our work adheres to the ACL Ethics Policy and uses publicly available datasets for reproducibility. LLMs may exhibit racial and gender biases, so we strongly recommend users assess potential biases before applying the models in specific contexts. Additionally, due to the difficulty of controlling LLM outputs, users should be cautious of issues arising from hallucinations.

\bibliography{custom}


\appendix

\section{Prompt Templates}
\label{app:prompts}

This appendix provides the full prompt templates used in each stage of RouterKGQA. Placeholders are shown in \texttt{<angle brackets>}.

\subsection{Stage 1: CRP Generation (specialized)}
\label{app:prompt_mrp}

The specialized model is fine-tuned on (question, gold CRP) pairs using the following instruction format:

\begin{tcolorbox}[colback=gray!5, colframe=gray!60, title=CRP Generation Prompt, breakable]
\small
\textbf{Instruction:} Generate a reasoning path that retrieves the information corresponding to the given question.

\textbf{Input:} Question: \{~\texttt{<question>}~\}

\textbf{Output:} \texttt{<structured CRP>}
\end{tcolorbox}

\subsection{Stage 2: Reasoning Blueprint Generation (general)}
\label{app:prompt_blueprint}

When the specialized-generated main path is unreachable on the KG, the general LLM decomposes the question into a sequence of single-hop reasoning steps via the following few-shot prompt:

\begin{tcolorbox}[colback=gray!5, colframe=gray!60, title=Reasoning Blueprint Prompt, breakable]
\small
Given a question, generate the reasoning steps without providing the final answer or specific entities.

Q: What countries are located in Eastern Europe and have the country calling code +373?\\
A: \#1 Identify the countries located in Eastern Europe.\\
\#2 Determine which of these countries has the country calling code +373.

Q: What country that trades with Turkey has an ISO numeric code lower than 012?\\
A: \#1 Identify the countries that are trade partners of Turkey.\\
\#2 Determine which of these countries has an ISO numeric code lower than 012.

Q: Who were the inspirations for the author of This Side of Paradise?\\
A: \#1 Identify the author of This Side of Paradise.\\
\#2 Determine the individuals or works that influenced the author.

Q: What countries border the location where the film Amen is set?\\
A: \#1 Identify the location where the film Amen is set.\\
\#2 Determine the countries that border this location.

Q: Lou Seal is the mascot for the team that last won the World Series when?\\
A: \#1 Identify the team for which Lou Seal is the mascot.\\
\#2 Determine the year this team last won the World Series.

Here is the question.\\
Q: \texttt{<question>}\\
A:
\end{tcolorbox}

\subsection{Stage 2: Path Selection}
\label{app:prompt_path_selection}

At each beam search step, after SBERT-based coarse filtering, the general LLM selects the most promising paths from the candidate set. The same prompt template is used for both intermediate steps (selecting up to $w$ paths) and the final step (selecting 1 path), with only the value of $n$ varying:

\begin{tcolorbox}[colback=gray!5, colframe=gray!60, title=Path Selection Prompt, breakable]
\small
Q: `\texttt{<question>}'\\
Paths from `\texttt{<start\_entity\_names>}':\\
Path 1: \texttt{<entity>} -> \texttt{<relation\_1>} -> \ldots{} -> \texttt{<end\_entity>}\ldots\\
Path 2: \texttt{<entity>} -> \texttt{<relation\_1>} -> \ldots{} -> \texttt{<end\_entity>}\ldots\\
\ldots

Select up to \texttt{<n>} paths most likely to reach the FINAL answer (not intermediate entities). Reply with path numbers only, e.g.: Path 1, Path 2
\end{tcolorbox}

\noindent During intermediate beam search steps, $n$ is set to the beam width $w$; at the final step (maximum depth reached), $n{=}1$ to select a single best path.

\section{SPARQL-to-CRP Conversion}
\label{app:mrp_conversion}

Gold CRPs used for specialized training are deterministically derived from
SPARQL annotations via the following rule-based procedure.

\paragraph{Step 1: Subgraph construction and main path extraction.}
We construct a directed graph from all triple patterns in the
\texttt{WHERE} clause, where each variable or named entity is a node and
each triple pattern $(s, r, o)$ forms a directed edge.
The topic entity $s_1$ is identified as the Freebase MID that lies on the
path to the \texttt{SELECT} target variable $o_N$.
We then extract the path from $s_1$ to $o_N$ in this graph;
the relation sequence along this path forms the main path
$\mathbf{R}=\langle r_1, \ldots, r_N \rangle$.
Since constraint branches attach to intermediate variables but do not
connect to $o_N$, this path is unique in the constructed graph.

\paragraph{Step 2: Constraint classification.}
All triple patterns and \texttt{FILTER} clauses not on the main path are
classified into three constraint types, each represented as
$C=(i, r_i^c, o_i^c)$:

\begin{itemize}
  \item \textbf{Entity restrictions.} Starting from each intermediate
    entity $o_i$ on the main path, we search over the remaining subgraph.
    Any branch that terminates at a named KG entity (i.e., a Freebase MID
    not on the main path) yields an entity restriction, where $o_i^c$ is
    the target entity.

  \item \textbf{Numeric restrictions.} Triples whose object is a numeric
    or temporal literal (e.g., \texttt{xsd:dateTime}, \texttt{xsd:float})
    yield a numeric restriction, where $o_i^c$ encodes both the comparison
    operator ($=$, $\geq$, $\leq$, $>$) and the threshold value, extracted
    from the associated \texttt{FILTER} clause. \texttt{ORDER BY} clauses
    with \texttt{LIMIT~1} are also converted to numeric restrictions with
    $o_i^c$ set to \texttt{argmax} or \texttt{argmin}.

  \item \textbf{String restrictions.} Triples whose object is a string
    literal (optionally with a language tag) yield a string restriction,
    where $o_i^c$ is the literal string value.
\end{itemize}

\paragraph{Round-trip verification.}
We verify the fidelity of this conversion by round-trip testing
(SPARQL $\to$ CRP $\to$ SPARQL): both directions use deterministic rules.

\section{Experimental Setup Details}
\label{app:experiment_setup}

\subsection{Dataset Details}
\label{app:dataset_details}

We evaluate RouterKGQA on two standard Freebase-based KBQA benchmarks: WebQuestionsSP (WebQSP)~\citep{yih-etal-2016-value} and Complex WebQuestions (CWQ)~\citep{talmor-berant-2018-web}. WebQSP contains 4,737 questions and is commonly used to evaluate KBQA systems on relatively simpler semantic parsing and multi-hop reasoning cases. CWQ is substantially larger, with 34,689 questions, and is designed to emphasize more complex compositional reasoning. Compared with WebQSP, CWQ contains a much larger proportion of deeper multi-hop questions and questions with richer constraints, making it a more challenging benchmark for constraint-aware KGQA. Table~\ref{tab:test_complexity_stats} reports the distributions of question depth and constraint complexity on the test sets of the two benchmarks.

\begin{table}[!htbp]
\centering
\small
\setlength{\tabcolsep}{4pt}
\begin{tabular}{llrrrr}
\toprule
\textbf{Aspect} & \textbf{Category} & \multicolumn{2}{c}{\textbf{CWQ}} & \multicolumn{2}{c}{\textbf{WebQSP}} \\
\cmidrule(lr){3-4}\cmidrule(lr){5-6}
 &  & \textbf{\#} & \textbf{\%} & \textbf{\#} & \textbf{\%} \\
\midrule
\multirow[c]{3}{*}{Depth}
& 1-hop   & 854  & 24.19 & 1,042 & 63.58 \\
& 2-hop   & 1,859 & 52.65 & 592   & 36.12 \\
& 3-hop+  & 818  & 23.17 & 5     & 0.31 \\
\midrule
\multirow[c]{4}{*}{Constraints}
& 0       & 1,176 & 33.31 & 1,169 & 71.32 \\
& 1       & 1,511 & 42.79 & 332   & 20.26 \\
& 2       & 633   & 17.92 & 92    & 5.61 \\
& 3+      & 211   & 5.98  & 46    & 2.81 \\
\bottomrule
\end{tabular}
\caption{Test-set distributions of question depth and constraint complexity on WebQSP and CWQ.}
\label{tab:test_complexity_stats}
\end{table}

\subsection{Metric Definitions}
\label{app:metric_details}

Prior KBQA methods adopt inconsistent definitions of Hits@1, which complicates fair comparison across approaches. We identify two answer-retrieval paradigms and describe how each has computed this metric.

\paragraph{Query-based methods.}
These methods retrieve answers by executing a structured query (e.g., SPARQL) against the KG. The returned answer set is inherently \emph{unordered}. ChatKBQA~\citep{luo-etal-2024-chatkbqa} counts Hits@1${=}1$ if the answer set of the first executable predicted query intersects with the gold answer set. In contrast, MemQ~\citep{xu-etal-2025-memory} considers only the \emph{first} returned entity---whose position is determined by the query engine and does not carry ranking semantics---and checks whether it belongs to the gold answer set.

\paragraph{LLM-filtering methods.}
These methods use an LLM to select or generate answers from candidate entities. Both RoG~\citep{luo2024rog} and GCR~\citep{pmlr-v267-luo25t} compute the metric identically: checking whether the LLM-filtered answer set has a non-empty intersection with the gold answer set. However, RoG labels this metric Hits@1, while GCR refers to it as Hit, despite the underlying computation being the same.

\paragraph{Unified evaluation protocol.}
To ensure a fair comparison across methods, we standardize the computation as follows:
\begin{itemize}
    \item For \textbf{query-based methods}, Hits@1${=}1$ if the answer set returned by the \emph{first executable query} has a non-empty intersection with the gold answer set.
    \item For \textbf{LLM-filtering methods}, Hits@1${=}1$ if the answer set generated by the \emph{first LLM reasoning step} has a non-empty intersection with the gold answer set.
\end{itemize}
F1 is computed using standard set-level precision and recall between the predicted and gold answer sets. This unified protocol ensures that Hits@1 reflects answer correctness rather than implementation-specific ordering effects.

\section{Implementation Details}
\label{app:implementation}

\subsection{Training Details}

The specialized CRP generation model is fine-tuned using LLaMA-Factory\footnote{\url{https://github.com/hiyouga/LLaMA-Factory}} with LoRA~\citep{hu2022lora}. We train dataset-specific specialized models for evaluation on WebQSP and CWQ. We apply the same training recipe to both Llama-2-7B and Llama-3.1-8B backbones unless otherwise specified. The main optimization settings are shared across datasets, while the number of training epochs differs between WebQSP and CWQ. Table~\ref{tab:hyperparams} summarizes the main training hyperparameters.

\begin{table}[h]
\centering
\small
\setlength{\tabcolsep}{4pt}
\begin{tabular}{lcc}
\toprule
\textbf{Hyperparameter} & \textbf{WebQSP} & \textbf{CWQ} \\
\midrule
\multicolumn{3}{l}{\emph{Stage 1: CRP Generation (Specialized Model)}} \\
Fine-tuning Type & LoRA & LoRA \\
Learning Rate & 5e-5 & 5e-5 \\
LR Scheduler & Cosine & Cosine \\
Train Batch Size & 4 & 4 \\
Gradient Accum. Steps & 4 & 4 \\
Effective Batch Size & 16 & 16 \\
Train Epochs & 50 & 10 \\
Precision & bf16 & bf16 \\
\midrule
\multicolumn{3}{l}{\emph{Stage 2: Path Repair (General Model)}} \\
General Model & GPT-4o-mini & GPT-4o-mini \\
Temperature $T$ & 0 & 0 \\
Beam Width $w$ & 3 & 3 \\
Relation Filter Size $x$ & 4 & 4 \\
Path Filter Size $y$ & 10 & 10 \\
Semantic Matching & \multicolumn{2}{c}{nomic-embed-text-v1} \\
\midrule
\multicolumn{3}{l}{\emph{Stage 3: Progressive Query Execution}} \\
Relaxation Order & \multicolumn{2}{c}{string $\to$ numeric $\to$ entity} \\
\bottomrule
\end{tabular}
\caption{Implementation details of RouterKGQA. Stage~1 settings are shared across Llama-2-7B and Llama-3.1-8B unless otherwise noted.}
\label{tab:hyperparams}
\end{table}

\subsection{Baseline Reproduction}

Table~\ref{tab:baseline_reprod} summarizes the reproduction status of all compared methods. For methods with publicly available code, we reproduce results in our environment using the same Freebase snapshot to ensure fair comparison. All reproduced methods follow the original paper's settings unless otherwise noted. For methods without public code, we report results directly from the original papers.

\begin{table}[h]
\centering
\small
\setlength{\tabcolsep}{3pt}
\scalebox{0.95}{
\begin{tabular}{llcl}
\toprule
\textbf{Category} & \textbf{Method} & \textbf{Reprod.} & \textbf{Notes} \\
\midrule
\multirow{3}{*}{LLM-only}
& Llama-2-7B       & \ding{51} & Direct inference \\
& Llama-3.1-8B     & \ding{51} & Direct inference \\
& GPT-4o-mini      & \ding{51} & Direct inference \\
\midrule
\multirow{6}{*}{IR-based}
& KV-Mem             & \ding{55} & Original paper \\
& PullNet            & \ding{55} & Original paper \\
& EmbedKGQA          & \ding{55} & Original paper \\
& NSM+h              & \ding{55} & Original paper \\
& TransferNet        & \ding{55} & Original paper \\
& Subgraph Retrieval & \ding{55} & Original paper \\
\midrule
\multirow{9}{*}{LLM-based}
& ToG              & \ding{51} & GPT-4o-mini \\
& PoG              & \ding{51} & GPT-4o-mini \\
& InteractiveKBQA  & \ding{55} & Original paper \\
& EffiQA           & \ding{55} & Original paper \\
& RoG              & \ding{51} & Official code \\
& ChatKBQA         & \ding{51} & Modified\textsuperscript{$\dagger$} \\
& MemQ             & \ding{51} & Official code \\
& GNN-RAG          & \ding{51} & Official code \\
& GCR              & \ding{51} & Official code \\
\bottomrule
\end{tabular}
}
\caption{Baseline reproduction status. \ding{51}: reproduced in our environment; \ding{55}: results from the original paper. \textsuperscript{$\dagger$}The original ChatKBQA uses Llama-2-13B on CWQ and Llama-2-7B on WebQSP; we unify both to Llama-2-7B for fair comparison under the same model scale.}
\label{tab:baseline_reprod}
\end{table}

\section{Sub-Question Consistency Analysis}
\label{app:subq_consistency}

To assess the semantic alignment between predicted reasoning paths and the original question, we reconstruct each predicted path into a natural-language question $q_r$ using an LLM and decompose both $q_r$ and the original question $q$ into sub-question sets, denoted by $A$ and $B$, respectively, following the decomposition style of CWQ~\citep{talmor-berant-2018-web}. Based on the set relationship between $A$ and $B$, we classify each example into four categories: \textsc{EXACT\_EQUAL} ($A=B$), \textsc{RECON\_SUBSET} ($A\subset B$), \textsc{PARTIAL\_OVERLAP} ($A\cap B\neq\emptyset$, with neither set containing the other), and \textsc{DISJOINT} ($A\cap B=\emptyset$).

\begin{figure*}[t]
\centering
\includegraphics[width=\textwidth]{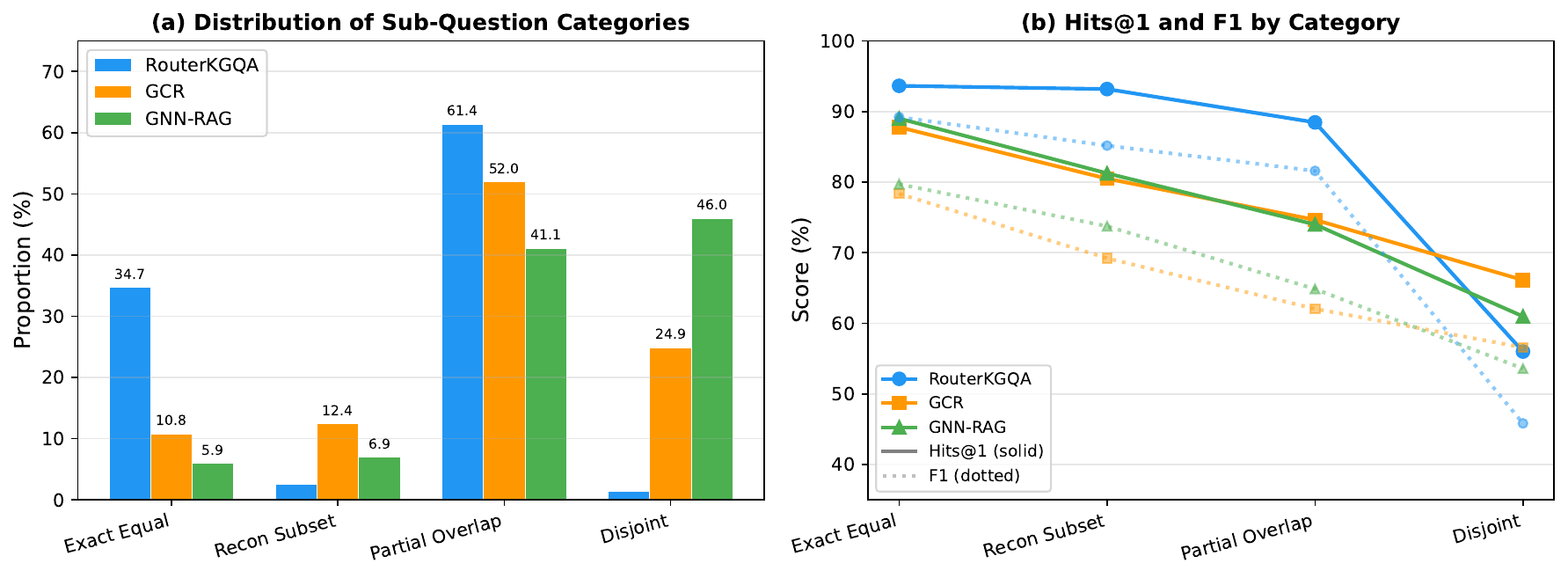}
\caption{Sub-question consistency analysis on CWQ. (a) Distribution of sub-question set relationship categories across methods. (b) Hits@1 (solid) and F1 (dotted) by sub-question relationship category.}
\label{fig:subq_combined}
\end{figure*}

Figure~\ref{fig:subq_combined}(b) confirms that semantic alignment directly determines downstream performance: \textsc{EXACT\_EQUAL} yields the best results across all methods, while \textsc{DISJOINT} yields the worst. Figure~\ref{fig:subq_combined}(a) shows that RouterKGQA substantially increases the proportion of \textsc{EXACT\_EQUAL} predictions (34.7\% vs.\ 10.8\% for GCR and 5.9\% for GNN-RAG) while reducing \textsc{DISJOINT} to 1.4\% (vs.\ 24.9\% and 46.0\%). This demonstrates that RouterKGQA generates reasoning paths that are semantically equivalent or highly consistent with the original question intent.

\clearpage

\end{document}